\documentclass[conference]{IEEEtran}

\usepackage{textpos}

\usepackage[pdftex]{graphicx}
\usepackage{amsmath}
\usepackage{algorithmic}
\usepackage{url}
\usepackage{amsmath,amsthm,amssymb,bm}
\usepackage[linesnumbered, ruled]{algorithm2e}

\SetCommentSty{mycommfont}

\begin{document}
%
\title{Phase Transition Adaptation}

\author{\IEEEauthorblockN{Claudio Gallicchio}
\IEEEauthorblockA{Department of Computer Science\\
University of Pisa, Italy\\
gallicch@di.unipi.it}
\and
\IEEEauthorblockN{Alessio Micheli}
\IEEEauthorblockA{Department of Computer Science\\
University of Pisa, Italy\\
micheli@di.unipi.it}
\and
\IEEEauthorblockN{Luca Silvestri}
\IEEEauthorblockA{silvestriluca@hotmail.it}}

\maketitle

\begin{textblock*}{150mm}(-1cm,-6cm)
\emph{This is a pre-print of a paper accepted at IJCNN-21.}
\end{textblock*}

\begin{abstract}
Artificial Recurrent Neural Networks are a powerful information processing abstraction, and Reservoir Computing provides an efficient strategy to build robust implementations by projecting external inputs into high dimensional dynamical system trajectories. In this paper, we propose an extension of the original approach, a local unsupervised learning mechanism we call Phase Transition Adaptation, designed to drive the system dynamics towards the `edge of stability'. Here, the complex behavior exhibited by the system elicits an enhancement in its overall computational capacity. We show experimentally that our approach consistently achieves its purpose over several datasets.
\end{abstract}

\IEEEpeerreviewmaketitle

\section{Introduction}
Recurrent Neural Networks (RNNs) \cite{rumelhart1986} present themselves as a powerful and popular bio-inspired approach that can learn relationships between time-varying input and output with complex spatiotemporal dependencies. Consisting of a brain-inspired network of processing units recurrently connected, they have been proved able of universal computation \cite{siegelmann1991, siegelmann1995}, and many different kinds of 'flavours' have been reported, e.g. \cite{hopfield1982, jordan1986, elman1990}. As it often happens in Machine Learning, standard RNNs need to go through a training phase in order to achieve good performance; and, as for feed-forward networks, the most popular training algorithms, i.e. Back-Propagation Through Time (BPTT) \cite{werbos1990} and Real Time Recurrent Learning (RTRL) \cite{williams1989}, are based on gradient descent techniques. 

However, these methods face several challenges: most notably, convergence, i.e. the movement of free parameters towards the optimal configuration, is slow - hampered by the presence of bifurcations in the parameters space, i.e qualitative changes in a system dynamics due to the modification of its parameters - and long-term dependencies in the data, i.e. patterns spanning inputs far in time, are difficult to be learned, due to the problem of vanishing/exploding gradients \cite{bengio1994}, i.e. computation of the gradient returns values either too small or too big. For these reasons, many alternatives have been explored to unlock RNNs full potential: from second-order methods, e.g. \cite{martens2011}, to evolutionary ones, e.g. \cite{schmidhuber2007, gomez2008}, along with architectural variants, e.g. Long-Short Term Memory (LSTM) \cite{hochreiter1997} and Gated Recurrent Unit (GRU) \cite{cho2014} - which introduce gated skip connections between the internal states of the model at subsequent time-steps to help errors flow backward in time.

Among these proposals emerged a different paradigm, called Reservoir Computing (RC) \cite{lukosevicius2009}, originally and independently introduced by \cite{jaeger2001}, under the name of Echo State Network (ESN), and by \cite{maass2002}, under the name of Liquid State Machine.
The main idea behind it consists in decoupling the system functions in two main components, each responsible for one: a non-linear high-dimensional dynamical system, called the \emph{reservoir}, and a trained linear readout. The first component is (usually) randomly initialized and thereafter left untrained, thus saving the expensive cost associated to the training of recurrent connections; its aim is to project the input stream unto a high-dimensional trajectory of transient states \cite{rabinovich2008}, where separation among different samples is more probable \cite{cover1965}.
The reservoir then feeds these transient projections to the second component - whose parameters are tuned through a supervised learning algorithm to minimize prediction error - to transform them into a stable target output.

Yet, a reservoir random initialization is unsatisfactory: not only space for optimization is left unexplored, but high fluctuations in performance are often observable, even when the same set of hyper-parameters describing the system, such as spectral radius and input scaling, is adopted \cite{ozturk2007}. Two main research directions - which would also promote efficient hardware realizations, ranging from electronic to optical solutions \cite{tanaka2019recent,brunner2013, larger2012, paquot2012} - tackle this drawback: the establishment of sound reservoir initialization strategies, topologically speaking as well as in terms of weights calibration \cite{rodan2011, appeltant2011, strauss2012, appeltant2014, ozturk2007}, and the development of adaptation strategies targeting useful system properties, e.g. reservoir state entropy \cite{triesch2005}.

In both cases, characterizations of dynamical systems highlighting properties associated to increased computational performance are highly desirable. Concerning this point, while looking for the conditions that support the emergence of computation in physical system, \cite{langton1990} observes that cellular automata, a class of formal abstractions of physical systems, exhibit the most complex behaviour in the vicinity of a phase transition between ordered and chaotic dynamics; there, the greatest potential to support information storage, transfer and modification is achieved. These considerations also align interestingly with theories about self-organizing criticality in extended dissipative coupled dynamical systems \cite{bak1988}, life included \cite{kauffman1993}.

In this work, elaborating over these observations, we will design an unsupervised adaptation mechanism, called Phase Transition Adaptation (PTA). Leveraging the property of a specific reservoir topology, this will enable us to drive system dynamics towards the 'edge of stability', a critical region where optimal computation in RNNs is expected \cite{busing2010, bertschinger2004, snyder2013}. 
Furthermore, such an algorithm, being local and cheap with respect to the state computation cost, may be a candidate mechanism for real-world neural networks adaptation.

This paper is organized as follows. We present the relevant background on RC/ESNs in Section~\ref{sec.ESN}. In the same section we also introduce the reader to the concepts of edge of stability and reservoir topology. These are central for the development of our proposed training algorithm, PTA, described in Section~\ref{sec.PTA}. 
Then, in Section~\ref{sec.experiments} we experimentally assess our methodology in comparison to related ESN
models. Finally, Section~\ref{sec.conclusions} concludes the paper.

\section{Echo State Networks}
\label{sec.ESN}
When dealing with temporally structured (sequential) inputs, NNs must be able to capture the (possible) dependence of a given target element $\mathbf{d}(t)$ on previous input elements $ \ldots, \mathbf{u}(t - 2), \mathbf{u}(t - 1)$, as well as on current input $\mathbf{u}(t)$.
Recurrent Neural Networks (RNNs) achieve this goal by representing past network inputs implicitly, both spatially and temporally, inside a network state, coinciding with the activation of its neurons in the recurrent hidden layer.
Training RNNs conventionally requires the calibration of all of its free parameters, i.e. all the weights of the input, recurrent and output connections. Following the success of the Back-Propagation (BP) algorithm to train standard NNs \cite{rumelhart1986}, gradient based methods to achieve this task have been designed, most notably Back-Propagation Through Time (BPTT) \cite{werbos1990} and Real Time Recurrent Learning (RTRL) \cite{williams1989}. 

However, both of them incur in serious problems related to the recurrent dynamics characterizing the network, which is an input-driven dynamical system. Qualitative changes in the system dynamics, e.g. the appearance of new attractors and/or the disappearance of existing ones, make the traditional BP error function quite complex, with lots of local minima, and vanishing or exploding gradients are frequent, i.e. computation of the gradient often brings up results either going to zero or blowing up, thus hampering training. Along a wide range of techniques designed to lessen these problems, radical new approaches have been proposed. Among these, our interest focuses on Echo State Networks (ESNs) \cite{jaeger2004,jaeger2001}, a Reservoir Computing (RC) instance that avoids to train recurrent weights while still providing state-of-the-art performance over many tasks.

\begin{figure}[t]
\centering
\includegraphics[width=\columnwidth]{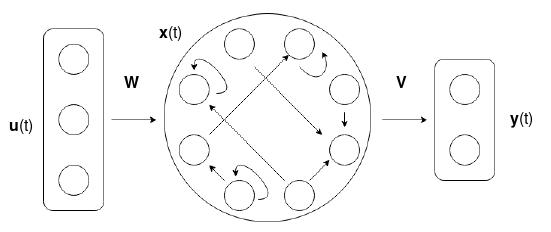}
\caption{Echo State Network. The network consists of an input layer (left), a reservoir (middle), and a readout layer (right).}
\label{fig:esn}
\end{figure}

Let $\mathbf{u}(t) \in \mathbb{R}^{U}$, $\mathbf{x}(t) \in \mathbb{R}^{N}$ and $\mathbf{y}(t) \in \mathbb{R}^{Y}$ be the external input, reservoir state and output of the system at time $t$, respectively.
An ESN state transfer function has the following form:
\begin{equation}\label{state_transfer_function}
\mathbf{x}(t) = f(\mathbf{x}(t - 1), \mathbf{u}(t)) = f(\hat{\mathbf{W}} \mathbf{x}(t - 1) + \mathbf{W} \mathbf{u}(t) + \bm{b}),
\end{equation}
while for the readout
\begin{equation}
\mathbf{y}(t) = g(\mathbf{V}\mathbf{x}(t)),
\end{equation}
where $\hat{\mathbf{W}} \in \mathbb{R}^{N \times N}$ is the recurrent weight matrix, 
$\mathbf{W} \in \mathbb{R}^{N \times U}$ is the input matrix, $\bm{b}$ is a bias vector,
$f$ is an element-wise non-linear sigmoidal function, $\mathbf{V} \in \mathbb{R}^{Y \times N}$ is the reservoir-to-output matrix and $g$ is an element-wise readout function (see Fig.\ref{fig:esn} for a graphical depiction). Following, we assume $f$ to be the hyperbolic tangent $\tanh$ and $g$ the identity function. 

The only matrix subject to training is $\mathbf{V}$, while $\mathbf{W}$ and $\hat{\mathbf{W}}$ are randomly initialized under stability criteria and then kept fixed. In particular, initialization of the untrained matrices needs to be consistent with the so called `echo state property' (ESP) \cite{jaeger2001}, which guarantees that the input-driven network's dynamics has a unique global attractor. 
A common strategy to perform the initialization of $\mathbf{\hat{W}}$ consists in three simple steps. First, initialize $\mathbf{\hat{W}}$ randomly (e.g. with weights distributed uniformly in $[-1, 1]$). Then, compute its spectral radius $\hat \rho$ (i.e., its largest eigenvalue in modulus). Finally, re-scale $\mathbf{\hat{W}} = \rho\; (\mathbf{\hat{W}}/{\hat \rho})$ to a desired value of the spectral radius $\rho$, exploring values below unity for a stable initialization of the dynamics in practice \cite{jaeger2001}. The input weight matrix $\mathbf{W}$ is initialized similarly, by first randomly drawing its elements from a uniform distribution on $[-1, 1]$, and then re-scaling them by an input scaling coefficient $\omega$. 
Finally, $\bm{b}$ is initialized in the same fashion as $\mathbf{W}$, and it is re-scaled by a bias scaling coefficient $\omega_b$.

Matrix $\mathbf{V}$ is typically found in closed-form, by exploiting linear methods such as Ridge Regression. In particular, 
$\mathbf{V} = \mathbf{Y} \mathbf{X}^T (\mathbf{X} \mathbf{X}^T + \kappa \mathbf{I})^{-1}$,
where $\mathbf{Y}$ and $\mathbf{X}$ are column-wise concatenations of the target and state values (respectively) computed over a training set, 
$\mathbf{I}$ is the identity matrix, and 
$\kappa$ is a regularization hyper-parameter.
Typically, when collecting the state vectors for the purposes of training, a number of $\tau$ time-steps are discarded, to wash out the initial transient in which the dynamics can be still affected by the initial conditions.

\noindent
\textbf{Edge of Stability} -
Dynamical systems, like reservoir layers, may exhibit a wide variety of behaviours, from stable to chaotic. In the following, a system is said to be chaotic if arbitrary small state perturbations at a given instant will affect the state of the system long afterwards, rather then vanishing away; and stable otherwise \cite{strogatz1994}. The region in the parameter space associated to a bifurcation between stability and chaos is called the \textit{edge of stability} (or, the \emph{the edge of chaos}). Systems operating at this phase transition present interesting dynamics.
In particular, ESNs were observed in multiple occasions \cite{verstraeten2007, verstraeten2009, boedecker2012, livi2018} to achieve the highest information processing capability when configured on the edge of stability, resulting in high memory capacity and good performance.
The edge of stability, then, is an interesting dynamical regime from a computational point of view. 

The edge of stability can be empirically detected by measuring Lyapunov exponents \cite{legenstein2005, verstraeten2007, verstraeten2009}. 
These
express
a measure of the asymptotic stability of autonomous dynamical systems, i.e. they describe the limit behaviour of a system as either chaotic or stable.
More formally, given an initial condition $\mathbf{x}_0$ and a perturbation $\delta_0$, trajectories diverge at a rate $||\delta_n|| \approx ||\delta_0|| e^{\lambda n}$,  where $\lambda$ is the Lyapunov exponent, and a positive value signals chaotic behaviour. 
If interested in the local behaviour of a system around any point $\mathbf{x}$ in phase space, useful metrics are the \textit{Local Lyapunov Exponents} (LLEs) \cite{bailey1996}, $\lambda = \log | \text{eig}(\mathbf{J}(\mathbf{x})) |$, where $\mathbf{J}(\mathbf{x})$ is the Jacobian\footnote{Given a function $f:\mathbb{R}^n \rightarrow \mathbb{R}^m$, where $f(\mathbf{x}) = [f_1(\mathbf{x}), \ldots, f_m(\mathbf{x})]$, its Jacobian $J \in \mathbb{R}^{m \times n}$ is defined as $J_{i,j} = \frac{\partial f_i(\mathbf{x})}{\partial \mathbf{x}_j}$.} matrix of the state transfer function computed in $\mathbf{x}$, and $\text{eig}(\mathbf{J}(\mathbf{x}))$ denotes its eigenvalues. Such approach coincides with a study of a system dynamical behaviour through its linearization around a given point 
\cite{verstraeten2009}. 
LLEs are well-defined only for autonomous systems, and thus, in case of ESNs, it would be more appropriate to talk about pseudo LLEs; still, we will continue to refer to them as LLEs. Averaging such quantities over many points of a given trajectory allows a fair guess at the system behaviour. %
Following \cite{bianchi2018}, we obtain the spectrum of LLEs as:
\begin{equation}
\lambda_k = \frac{1}{T} \sum_{t = 1}^T \log\left|\text{eig}_k(\mathbf{J}(t))\right|,
\end{equation}
where $\text{eig}_k(\mathbf{J}(t))$ is the k-th eigenvalue of the Jacobian computed at time $t$.\\

\noindent
\textbf{Reservoir Topology} - 
A profitable strand of works in RC research focuses on studying the topology of the reservoir layer. The idea is to try to generate ``better'' reservoirs than just random ones. An effective approach is to design the recurrent weight matrix $\mathbf{\hat W}$ as an orthogonal matrix, which has been shown to be associated to increased short-term memory and predictive performance with respect to random reservoirs \cite{white2004}.

\begin{figure}[t]
\centering
\includegraphics[width=\columnwidth]{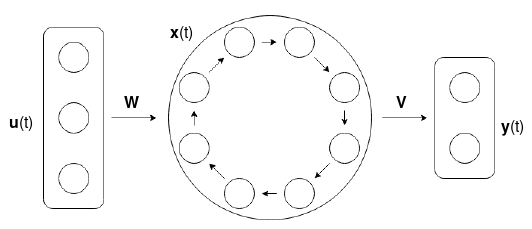}
\caption{Echo State Network with ring topology.}
\label{fig:ring}
\end{figure}

A particular sub-class of these matrices is that describing rings 
\cite{strauss2012}, graphs consisting in a single cycle (see Fig. \ref{fig:ring}). In an attempt to achieve minimal complexity for the reservoir topology and parameterization while retaining good performance levels, \cite{rodan2011} proposed a \textit{Simple Cycle Reservoir} (SCR) model, adopting this particular topology. In this paper we use SCR as a representative for this kind of ring-constrained and simplified architectures.
In particular, the non-zero weights in $\mathbf{\hat W}$ are limited to the elements in the sub-diagonal, and in the top right, and they are all set to the same value, i.e.:
\begin{equation}
\label{eq:scr_w}
\hat{\mathbf{W}} = 
\left(\begin{matrix}
     0 &      0 & \ldots &        0 &      \hat w\\
     \hat w &      0 & \ldots &        0 &      0\\
     0 & \ddots & \ddots &  \vdots  & \vdots\\
\vdots & \ddots & \ddots &        0 &      0\\
     0 & \ldots &      0 &        \hat w &      0
\end{matrix}\right).
\end{equation}
Given the particular reservoir topology, from the above \eqref{eq:scr_w}, we notice that $\rho = \hat w$. We can take advantage of this simplified eigenvalues computation to build our algorithm, described in the next Section~\ref{sec.PTA}.

\section{Phase Transition Adaptation}
\label{sec.PTA}
Here we introduce our local unsupervised adaptation algorithm, Phase Transition Adaptation (PTA), designed to drive system dynamics towards the edge of stability. We chose as our base architecture the SCR described in the previous Section~\ref{sec.ESN}, where the reservoir nodes are connected to form a ring. We modify it by introducing in the state transfer function two vectors of free parameters: a gain $\bm{a} \in \mathbb{R}^N$, to modulate the net input of the state transfer function, and an adaptable bias $\bm{b} \in \mathbb{R}^N$, to move its operational regime between linear and nonlinear. The reservoir weight $\hat w$ (see \eqref{eq:scr_w} is chosen to be $1$ such that gain tuning may incidentally corresponds with reservoir spectral radius tuning. We then re-formulate the reservoir state transfer function as follows:
\begin{equation}\label{eq:pta_state}
\mathbf{x}(t) = \tanh(\bm{a} \odot ({\hat{\mathbf{W}}} \mathbf{x}(t - 1) + \mathbf{W} \mathbf{u}(t)) + \bm{b}).
\end{equation}

\noindent
\textbf{Local Lyapunov Exponents of SCR} - In order to control the proximity of the system behaviour to the edge of stability, we want to find a suitable way of computing its LLEs, which in turns requires the computation of the Jacobian matrix of its state transfer function. In our case, the Jacobian $\mathbf{J}(t)$ of \eqref{eq:pta_state} 
is given by:
\begin{equation}
\label{eq.jacobian}
\mathbf{J}_{i,j}(t) = \left\{\begin{matrix}
 (1 - \mathbf{x}_i(t)^2) \bm{a}_i  &  \text{if } j = i - 1 \text{ or } (i,j) = (1,N)\\
0  & \quad \text{otherwise}
\end{matrix}\right.
\end{equation}
Notice that, interestingly, the Jacobian matrix retains the same simplified structure of the SCR reservoir weight matrix $\hat{\mathbf{W}}$, i.e. a ring-shaped topology.
Next, we need to retrieve the eigenvalues of $\mathbf{J}(t)$: to do this we must first compute its characteristic polynomial
$p_J(\gamma) = \det(\mathbf{J} - \gamma \mathbf{I}).$
Again, leveraging the ring structure of the matrix $\mathbf{J}(t)$, computation of $p_J(\gamma)$ through Laplace expansion is much simplified, and we obtain
$p_J(\gamma) = - \gamma^N + \prod_{k = 1}^N  \left( 1 + ((1 - \mathbf{x}_k(t)^2) \bm{a}_k - 1) \right)$.
Denoting $\bm{\eta}_k(t) = 1 + ((1 - \mathbf{x}_k(t)^2) \bm{a}_k - 1)$, we can then express the eigenvalues of $\mathbf{J}(t)$ as
$\gamma_j(t) = \left( \prod_{k = 1}^N \bm{\eta}_k(t) \right) ^ \frac{1}{N} e^\frac{i 2\pi j}{N} \quad \forall j \in [1,\ldots,N]$.
We observe that $| \gamma_i(t) | = | \gamma_j(t) | \; \forall i, j$, i.e. the eigenvalues of $\mathbf{J}(t)$ have all the same module; therefore, the LLEs associated to the network are all equal, i.e. $\lambda_i(t) = \lambda_j(t)$ $\forall i,j \in [1, \ldots, N]$. Thus, we write:
\begin{equation}\label{eq:lle_scr}
\lambda(t) = \log \left| \left( \prod_{k = 1}^N \bm{\eta}_k(t)  \right) ^ {\frac{1}{N}} \right| = 
\frac{1}{N} \sum_{k=1}^N \log \left| \bm{\eta}_k(t) \right|.
\end{equation}
With such a result, we move to define our algorithm update rule.

\noindent
\textbf{Update Rule Derivation} -
In order to push the dynamics of the reservoir towards the edge of stability, we need to drive the values of the Lyapunov exponents towards zero, i.e. we need to minimize 
$
e(t) = || \lambda(t) || ^ 2 = \sum_{i = 1}^N \lambda_i(t)^2 = 
\frac{1}{N} \left( \sum_{i = k}^N \log \left| \bm{\eta}_k(t) \right| \right)^2.
$
To do this, we 
use
Stochastic Gradient Descent (SGD) with momentum; thus, we first compute
the gradient values with respect to $\bm{a}$ and $\bm{b}$. Denoting by $\mathbf{net}(t)$ the value ${\hat{\mathbf{W}}} \mathbf{x}(t - 1) + \mathbf{W} \mathbf{u}(t)$, and using the subscript $i$ to indicate a quantity that is referred to the $i$-th reservoir neuron, we have:

\begin{equation}
\begin{split}
\frac{\partial e(t)}{\partial \bm{a}_i}
& =
\frac{2}{N}  \sum_k^N \log (\bm{\eta}_k(t)) \frac{\partial \log(\bm{\eta}_i(t))}{\partial \bm{a}_i}
=
2 \frac{\lambda(t)}{\bm{\eta}_i(t)} \frac{\partial \bm{\eta}_i(t)}{\partial \bm{a}_i} \\
& =
2 \frac{\lambda(t)}{\bm{\eta}_i(t)}  \left( (1 - \mathbf{x}_i(t)^2) + 
\frac{\partial(1 - \mathbf{x}_i(t)^2)}{\partial \bm{a}_i} \bm{a}_i \right) \\
& =
2 \frac{\lambda(t)}{\bm{\eta}_i(t)}  \left( (1 - \mathbf{x}_i(t)^2) - 2\mathbf{x}_i(t) \frac{\partial \mathbf{x}_i(t)}
{\partial \bm{a}_i} \bm{a}_i \right) \\
& =
2 \frac{\lambda(t)}{\bm{\eta}_i(t)} \left( (1 - \mathbf{x}_i(t)^2) - 2\mathbf{x}_i(t) (1 - \mathbf{x}_i(t)^2) 
\textbf{net}_i(t) \bm{a}_i \right) \\
& =
2 \frac{\lambda(t)}{\bm{\eta}_i(t)} (1 - \mathbf{x}_i(t)^2) \left( 1 - 2 \mathbf{x}_i(t) \textbf{net}_i(t) \bm{a}_i \right), 
\end{split}
\end{equation}
\begin{equation}
\begin{split}
\frac{\partial e(t)}{\partial \bm{b}_i}
& =
\frac{2}{N}  \sum_k^N \log (\bm{\eta}_k(t)) \frac{\partial \log(\bm{\eta}_i(t))}{\partial \bm{b}_i}
=
2 \frac{\lambda(t)}{\bm{\eta}_i(t)} \frac{\partial \bm{\eta}_i(t)}{\partial \bm{b}_i} \\
& =
2 \frac{\lambda(t)}{\bm{\eta}_i(t)} \frac{\partial(1 - \mathbf{x}_i(t)^2)}{\partial \bm{b}_i} \bm{a}_i \\
& =
-4 \frac{\lambda(t)}{\bm{\eta}_i(t)} \mathbf{x}_i(t) \frac{\partial \mathbf{x}_i(t)}{\partial \bm{b}_i} \bm{a}_i \\
& =
-4 \frac{\lambda(t)}{\bm{\eta}_i(t)} \mathbf{x}_i(t) (1 - \mathbf{x}_i(t)^2) \bm{a}_i.
\end{split}
\end{equation}
We then build the update equations:
\begin{equation}
\label{eq:dadb}
\begin{array}{l}
\Delta \bm{a}_i(t) = \alpha \Delta \bm{a}_i(t - 1) + (1 - \alpha) \frac{\partial e(t)}{\partial \bm{a}_i},\\
\\
\Delta \bm{b}_i(t) = \alpha \Delta \bm{b}_i(t - 1) + (1 - \alpha) \frac{\partial e(t)}{\partial \bm{b}_i},
\end{array}
\end{equation}
where $\alpha \in [0, 1]$ is a momentum parameter, useful to accelerate convergence.
Finally, the update rules for $\bm{a}$ and $\bm{b}$ at time $t + 1$ are, respectively
\begin{equation}
\label{rule:ab}
\begin{array}{l}
\bm{a}_i(t + 1) = \bm{a}_i(t) - \eta \Delta \bm{a}_i(t),
\\
\\
\bm{b}_i(t + 1) = \bm{b}_i(t) - \eta \Delta \bm{b}_i(t),
\end{array}
\end{equation}
where $\eta$ is the algorithm learning rate.

\noindent
\textbf{PTA training algorithm} -
Based on the mathematical derivation illustrated so far, we are now ready to introduce our proposed PTA training algorithm. The aim is to adapt the gains and biases of the recurrent (hidden) neurons to have system dynamics closer to the edge of stability.
PTA is an unsupervised online training algorithm, i.e. the parameters are updated (without target supervision) at each input sample presented to the network, meaning that the network can adapt its parameters in an ongoing fashion during the exposition to external stimuli, and thus adjust its dynamics concurrently to the execution of a task.

The pseudo-code of the training algorithm is presented in Algorithm~\ref{code:pta}.
PTA takes a dataset containing an input time-series of length $T$, i.e., $\mathbf{u}(1), \mathbf{u}(2), \ldots, \mathbf{u}(T)$, and produces the final values for $\bm{a}$ and $\bm{b}$. The algorithm starts by initializing the reservoir, following the strategy delineated in Section~\ref{sec.ESN}, using a ring-shaped connectivity where the non-zero weights are all set to the value $\hat w = 1$ (see \eqref{eq:scr_w}). The components of the gain vector $\bm{a}$ are all set to an initial value 
$\rho$, and those of the bias vector $\bm{b}$ are set to a unitary initial value.
For every epoch, the algorithm first discards an initial transient of length $\tau$ (limiting itself to updating the reservoir state). Then, for each time-step after the transient, the algorithm updates the values of $\bm{a}$ and $\bm{b}$ (which requires the computation of the state and LLEs).
The procedure is iterated until a maximum number of epochs is reached, denoted as $MAX_{epochs}$, or until the value of $\lambda$ becomes greater then a tolerance threshold $\vartheta_{\lambda}$\footnote{We use this threshold to control the proximity to the edge of stability. To avoid numerical instabilities we fix this value to $\vartheta_{\lambda} = -0.1$.}.
The hyper-parameters requested by the algorithm are the learning rate $\eta$, the momentum $\alpha$, the maximum number of epochs $MAX_{epochs}$, and those necessary to initialize the reservoir, i.e., the input scaling $\omega$ and the initial value for the gains $\rho$, which plays a similar role to the spectral radius in common RC setups. Notice, however, that while in standard RC applications this value needs to be tuned by hyper-parameter search, in PTA the gain vector $\bm{a}$ is subject to training.
Finally, notice that the PTA algorithm given in Algorithm~\ref{code:pta} can be easily extended to cope with training sets comprising multiple time-series, by iterating the same process in lines 4-12 for every time-series.

\noindent
\textbf{Complexity Estimation} - 
For an estimation of the time complexity of the Algorithm~\ref{code:pta}, let $T$ be the length of the input sequence,, $N$ the number of units in the reservoir, and $U$ the input dimension. For each input sample (see lines $8$-$11$ in Algorithm~\ref{code:pta}), our algorithm computes reservoir state, Lyapunov exponent and update values. 
Looking at the corresponding equations, we conclude that its time complexity is the same of the matrix-vector product $\mathbf{W}\mathbf{u}(t)$, i.e. $\mathcal{O}(N \; U)$. Assuming to reiterate it for all input elements of the given input sequence
for a certain number of epochs $E$, 
the cost becomes $\mathcal{O}(N \; U \; T \; E)$, which scales linearly with the number of recurrent units and the length of the training sequence.

For a space complexity estimation, we observe that, since the computations 
involve only intermediate scalar values, and they must be repeated for each unit, the algorithm space complexity is $\mathcal{O}(N)$, which - again - scales linearly with the number of recurrent neurons.

\begin{algorithm}\label{code:pta}
\SetAlgoLined
\SetKwRepeat{Do}{do}{while}
 \KwData{Input time-series $\mathbf{u}(1), \mathbf{u}(2), \ldots$}
 \KwResult{Final values for gain $\bm{a}$ and bias $\bm{b}$}
 initialize the network\;
 initialize $epoch = 0$\;
 \Do{$epoch < MAX_{epochs}$ and $\lambda < \vartheta_{\lambda}$}{
  \tcp{wash out an initial transient}
  \For{$t = 1, 2, \ldots, \tau$}{
  compute $\mathbf{x}$\ using \eqref{eq:pta_state}\;
  }
  \For{$t = \tau+1, \tau +2, ..., T$}{
  compute $\mathbf{x}$\ using \eqref{eq:pta_state};\\
  compute $\lambda$\ using \eqref{eq:lle_scr};\\
  compute $\Delta \bm{a}$, $\Delta \bm{b}$
  using \eqref{eq:dadb}
  ;\\
  update($\bm{a}$, $\bm{b}$) using \eqref{rule:ab};
  }
  $epoch$++\;
  }
 \caption{Phase Transition Adaptation}
\end{algorithm}


\section{Experiments}
\label{sec.experiments}
In this section we report the details of the experimental analysis conducted in this paper. We first introduce the tasks in Section~\ref{sec.tasks}. Then, we describe the experimental settings used in our experiments, in Section~\ref{sec.settings}. Finally, in Section~\ref{sec.results} we report the outcomes of the experimental analysis.

\subsection{Tasks}
\label{sec.tasks}
We adopt a number of benchmarks tasks from the RC literature. Specifically, we first consider a direct measure of the short-term memorization abilities of the reservoir system. Then, we consider a number of tasks that exercise a combination of memorization and non-linear processing capabilities.
For all the datasets described in the following we constructed a time-series of length $20000$, where the first $15000$ time-steps were used for training and the remaining $5000$ for test. For hyper-parameter tuning we used a validation set comprising the last $5000$ time-steps of the training set.

\noindent
\textbf{Memory Capacity (MC)} - 
An important property of systems operating on time-series is their information storage capacity, i.e., their short-term memory. A measure of it denoted Memory Capacity (MC) was introduced in \cite{jaeger2002}, and it consists in assessing the network capability to recall progressively delayed input samples coming from an uni-dimensional input.
To maximally test the system's memory, each input element $u(t)$ is drawn randomly from a uniform distribution in $[0,0.5]$. The $k$-th readout neuron, whose output is denoted as $y_k(t)$, is trained to reconstruct the input with a delay $k$, i.e. $u(t-k)$. The goodness of the reconstruction is measured by a MC score, computed as:
\begin{equation}
\label{eq:mc}
MC = \sum_{k=1}^\infty(r^2(u(t-k),y_k(t))),
\end{equation}
where $(r^2(\cdot,\cdot))$ indicates the squared correlation.
Theoretical results in \cite{jaeger2002} tell us that the MC score decreases for increasing delays, and that, for an $N$-dimensional reservoir, $MC \leq N$. Hence, in practice, the summation in \eqref{eq:mc} can be performed up to a finite number of delay terms, which we chose to be equal to $2 N$.

\noindent
\textbf{Nonlinear Memorization (NLM)} -
Findings indicate a trade-off between memory capacity and nonlinear processing capability
\cite{verstraeten2010memory}.
Interestingly, as discussed in  \cite{dambre2012}, information is not lost inside non-linear networks but merely transformed in various non-linear ways. This implies that non-linear mappings with memory in dynamical systems are not only realizable with linear systems providing the memory followed by a non-linear function (see \cite{digregorio2018} for an example of heterogeneous reservoirs), but that it is possible to find a system that inherently implements the desired mappings.
To investigate the extent of these observations, we consider a nonlinear memorization task \cite{inubushi2017}, where the input $u(t)$ is uniformly distributed in $[0,1]$ while the target consists in: 
\begin{equation*}\label{eq:nlm}
d(t) = \sin(\nu u(t - \delta)),
\end{equation*}
where $\nu$ and $\delta$ are parameters governing the `strength' of the non-linearity and the `depth' of the required memory, respectively.
In this case we chose $\nu = \sqrt{2}$ and $\delta = 30$.

\noindent
\textbf{NARMA} -
Introduced in \cite{atiya2000}, the nonlinear auto-regressive moving average (NARMA) system is a discrete time system. The current output depends on both the input and the previous outputs. In general, modeling this system is difficult due to the non-linearity and long memory required.
We considered systems of order $20$, defined by the following relation:
\begin{align*}\label{eq:narma}
d(t + 1) =  & \tanh \big(0.3 d(t) + 0.05d(t)
\sum_{i = 0}^{19} d(t - i) +\\ &1.5u(t - 19)u(t) + 0.01\big),
\end{align*}
where $u(t)$ is the system input at time $t$ (an i.i.d stream of values generated uniformly from an interval $[0, 0.5]$), and $d(t)$ is the target at time $t$.

\noindent
\textbf{Mackey-Glass (MG)} - 
The Mackey-Glass (MG) time-series prediction is a common benchmark in the RC community. Such a system is described through a nonlinear time delay differential equation \cite{mackey1977}, i.e.:
\begin{equation*}\label{eq:mg}
\frac{du(t)}{dt} = \beta \frac{u(t - \delta)}{1 + u(t - \delta)^n} - \gamma u(t),
\end{equation*}
where $\beta,\, \gamma,\, \delta,\,$ and $n$ are real numbers. To make the system exhibit chaotic behaviour, we choose $\beta = 0.2$, $\gamma = 0.1$, $\delta = 30$ and $n = 10$. The time-series for this task is obtained by discretizing the above differential equation, and setting up a next-step prediction task, where the target at each time $t$ is the value of the MG time-series one step ahead, i.e. d(t) = u(t+1).


\subsection{Experimental Settings}
\label{sec.settings}
In our experiments with PTA, we used reservoirs containing $N = 100$ neurons. Unless otherwise stated, we initialized the input-reservoir connections using an input scaling $\omega = 0.1$. For the PTA training algorithm, we initialized the gains with $\rho = 0.5$, and used learning rate $\eta = 10^{-5}$, and momentum $\alpha = 0.9$. We fixed the maximum number of epochs $MAX_{epochs} = 50$, although, in practice, only a few epochs were needed.
As detailed in Section~\ref{sec.PTA}, PTA training is unsupervised, hence only the input time-series is used to adapt the reservoir dynamics. After that, the readout was trained by using Ridge Regression as in standard ESNs (see Section~\ref{sec.ESN}), with regularization coefficient $\kappa = 10^{-8}$. The achieved results are averaged (and std computed) over $20$ repetitions (to account for the randomization aspects in the initialization).

For the aims of a comparative analysis, we performed the same experiments using standard RC networks, namely ESNs with fully connected reservoirs, and SCRs. 
Notice that, while PTA automatically adapts the gains and the biases of the reservoir neurons based on the driving input, 
in ESNs and SCRs the spectral radius $\rho$ and the bias scaling $\omega_b$, are treated as hyper-parameters. We tuned these hyper-parameters on the validation set by using random search from a uniform distribution in $(0,1)$, unless otherwise stated.
To make a fair comparison with PTA, the number of explored configurations in the random search is set to a value that makes the required computational time approximately the same  required by the PTA experiment on the same task.

We ran our experiments on a MacBook Pro laptop equipped with a 2,8 GHz Quad-Core Intel Core i7 processor and 16 GB of RAM. The code used in our experiments is written in Python/NumPy and is made publicly available\footnote{https://github.com/gallicch/PhaseTransitionAdaptation}.

\subsection{Results}
\label{sec.results}
In the following we discuss the experimental results, 
starting with the outcomes on the MC task, and then moving to the predictive tasks, i.e. NLM, NARMA and MG.

\noindent
\textbf{Experiments on MC} -
The experiments performed on this task were conducted within a slightly modified setting with respect to the general description given in Section~\ref{sec.settings}. Specifically, as the dynamical short-term memorization is known to be influenced by the linearity regime of the reservoir \cite{jaeger2002}, which in turn is affected by the input scaling $\omega$, we test PTA in correspondence of three values of the input scaling, namely $\omega = 0.01, 0.1, 1$.

\begin{figure}[!t]
\centering
\includegraphics[width=\columnwidth]{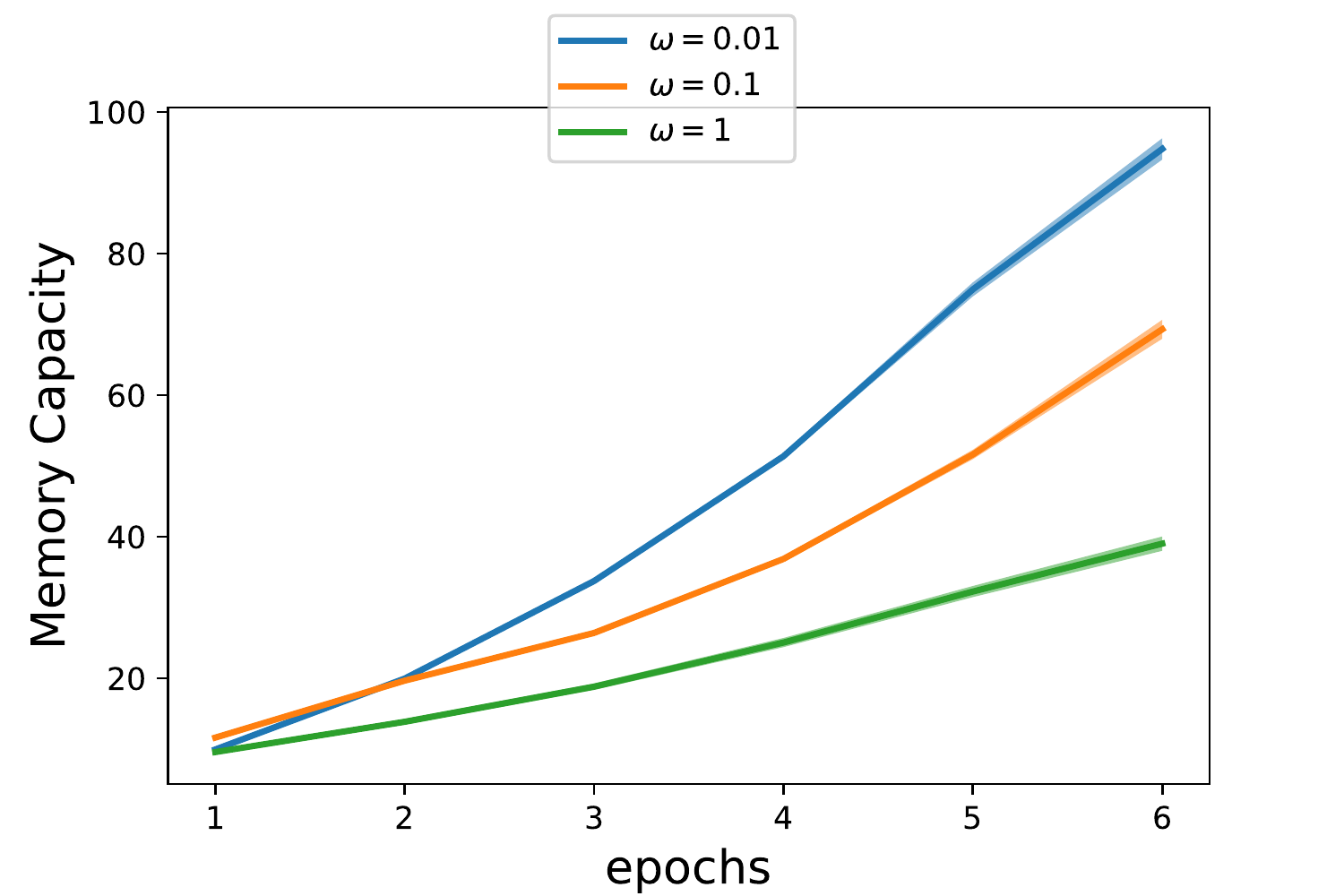}
\caption{Test MC score over the epochs of the PTA learning algorithm. The different lines in the plot refer to different values of the input scaling hyper-parameter $\omega$.}
\label{fig.MC}
\end{figure}

In Figure~\ref{fig.MC} we show the MC score measured on the test set after each epoch of PTA training. Results clearly indicate that
PTA progressively improves the MC over the epochs in all settings. This is not surprising, as MC is known to sensibly improve near the edge of stability. Yet, it shows the great practical efficacy of the proposed training algorithm to drive the reservoir dynamics towards such a behavior. 
The plot in Figure~\ref{fig.MC} also shows that, in all settings, training required no more than 6 epochs to achieve this result.
Moreover, in line with the known MC characterization, networks initialized with a smaller input scaling $\omega$ led to higher MC scores.
Overall, we can observe a substantial improvement, from a factor of $\approx 4$ (for $\omega = 1$) up to a factor of $\approx 9$ (for $\omega = 0.01$). 

Table~\ref{tab.results_mc} reports the MC score achieved on the test set after PTA training, comparatively with
ESN and SCR\footnote{In these experiments the upper bound for the bias scaling search is set to $\omega$, to uniformly treat the scaling coefficients in the input-reservoir connections.}.
Results show that PTA is definitely advantageous, reaching the highest MC in all settings. Interestingly, for $\omega = 0.01$, a MC score close to the theoretical limit (of 100) is obtained. Moreover, even for $\omega = 1$, the performance achieved by PTA is close to those of ESN and SCR initialized with state transitions much closer to the linear regime (respectively $\omega = 0.01$ and $\omega = 0.1$), where the memorization skills are potentially higher.

\begin{table*}[t]
\caption{MC score (the higher the better) on the test set achieved by ESN, SCR and PTA. Best results are highlighted in bold font.}
\label{tab.results_mc}
\normalsize
\centering
\begin{tabular}{llll}
\emph{Input Scaling}& \textbf{ESN} & \textbf{SCR} & \textbf{PTA}\\
\hline
\hline
$\omega = 1$ 
      & $16.12_{(\pm 0.50)}$
      & $19.99_{(\pm 0.69)}$ 
      & $\mathbf{41.07}_{(\pm 2.11)}$  \\
$\omega = 0.1$ 
   & $32.59 _{(\pm1.78)}$
   & $41.51 _{(\pm0.67)}$
   & $\mathbf{69.30 } _{(\pm1.34)}$ \\
$\omega = 0.01$ 
   & $42.76 _{(\pm3.72)}$
   & $93.02 _{(\pm0.80)}$
   & $\mathbf{94.78} _{(\pm1.50)}$ \\
\end{tabular}
\end{table*}

\noindent
\textbf{Predictive Tasks} -
In Table~\ref{tab.results_all} we report the Normalized Mean Squared Error (NMSE) obtained by PTA on NLM, NARMA and MG tasks. The same table provides the results achieved by ESN and SCR, for comparison. As it is evident, PTA outperforms both standard ESN and SCR on all the considered tasks. In particular, the error decreases by more than one order of magnitude in the case of NLM, and significant reductions (also considering the std) are obtained for NARMA and MG. 

Overall, the results shown in this section indicate the effectiveness of PTA in conducting the state dynamics in a particularly advantageous region, where the reservoir shows very good performance both on pure memorization and on predictive tasks that combine memorization requirements and non-linear processing capabilities.

\begin{table*}[tb]
\caption{NMSE (the lower the better) on the test set achieved by ESN, SCR and PTA. Best results are highlighted in bold font.}
\label{tab.results_all}
\normalsize
\centering
\begin{tabular}{llll}
\emph{Task}& \textbf{ESN} & \textbf{SCR} & \textbf{PTA}\\
\hline
\hline
NLM & $9.30 e-01 _{(\pm5.41 e-02)}$
   & $9.80 e-01 _{(\pm1.36 e-02)}$
   & $\mathbf{6.76 e-02} _{(\pm7.47 e-08)}$ \\
NARMA & $1.61 e-01_{(\pm 1.36 e-02)}$
      & $1.77 e-01_{(\pm 1.08 e-02)}$ 
      & $\mathbf{1.33 e-01}_{(\pm 3.04 e-03)}$  \\
MG & $9.31 e-06 _{(\pm1.61 e-07)}$
   & $7.14 e-07 _{(\pm5.73 e-08)}$
   & $\mathbf{5.90 e-07} _{(\pm6.35 e-08)}$ \\
\end{tabular}
\end{table*}

\section{Conclusion}
\label{sec.conclusions}
In this work we presented a novel local unsupervised training algorithm for Recurrent Neural Networks, called Phase Transition Adaptation (PTA). Our method manages to drive the behaviour of Simple Cycle Reservoirs (SRCs), restricted ring-topology Echo State Networks (ESNs), towards the 'edge of stability', a phase transition region between ordered and chaotic dynamics. We quantified such a behaviour looking at Local Lyapunov Exponents (LLEs), measures of local predictability in phase-space trajectories, and we developed a gradient descent approach to push the LLEs towards the critical zero value, leveraging the influence of specific network parameters - namely, gain and bias.

We assessed our algorithm over several datasets in comparison to both standard ESNs and SCRs. The reported results showed the ability of PTA to bring considerable performance enhancements both in terms of short-term memory, and in terms of predictive performance in tasks that fuse memorization with non-linear processing.

Thinking of further developments, it would be nice to enlarge our approach to encompass more complex reservoir-based neural networks, including Deep Reservoir Computing for temporal \cite{gallicchio2017desn} and graph data structures \cite{gallicchio2020fast}.
Moreover, the properties of being local, unsupervised and cheap, make the proposed PTA training algorithm a perfect candidate for implementing adaptation mechanisms in physical RNNs.
Also, we note that tuning the input scaling parameter is often fundamental to achieve a good performance, while early attempts to modify it with our algorithm did consistently zero it out. It would thus be desirable to find task-dependent input-scaling adaptation mechanisms to build synergies with our approach, as \cite{lazar2007} showed possible between intrinsic plasticity and spike timing dependent plasticity mechanisms.

In the end, the road we envision leads to a deeper understanding of the mechanisms underlying environmental adaptability in dynamical systems as means of computation, and, more ambitiously, to insights over the computational nature of our own cognitive adaptation strategies.


\bibliographystyle{IEEEtran}
\bibliography{references}

\begin{thebibliography}{10}
\providecommand{\url}[1]{#1}
\csname url@samestyle\endcsname
\providecommand{\newblock}{\relax}
\providecommand{\bibinfo}[2]{#2}
\providecommand{\BIBentrySTDinterwordspacing}{\spaceskip=0pt\relax}
\providecommand{\BIBentryALTinterwordstretchfactor}{4}
\providecommand{\BIBentryALTinterwordspacing}{\spaceskip=\fontdimen2\font plus
\BIBentryALTinterwordstretchfactor\fontdimen3\font minus
  \fontdimen4\font\relax}
\providecommand{\BIBforeignlanguage}[2]{{%
\expandafter\ifx\csname l@#1\endcsname\relax
\typeout{** WARNING: IEEEtran.bst: No hyphenation pattern has been}%
\typeout{** loaded for the language `#1'. Using the pattern for}%
\typeout{** the default language instead.}%
\else
\language=\csname l@#1\endcsname
\fi
#2}}
\providecommand{\BIBdecl}{\relax}
\BIBdecl

\bibitem{rumelhart1986}
D.~E. Rumelhart, G.~E. Hinton, and R.~J. Williams, ``Learning internal
  representations by error propagation,'' in \emph{Parallel Distributed
  Processing: Explorations in the Microstructure of Cognition, Vol. 1}.\hskip
  1em plus 0.5em minus 0.4em\relax Cambridge, MA, USA: MIT Press, 1986, pp.
  318--362.

\bibitem{siegelmann1991}
H.~T. Siegelmann and E.~D. Sontag, ``Turing computability with neural nets,''
  \emph{Applied Mathematics Letters}, vol.~4, no.~6, pp. 77--80, 1991.

\bibitem{siegelmann1995}
------, ``On the computational power of neural nets,'' \emph{Journal of
  Computer and System Sciences}, vol.~50, no.~1, pp. 132--150, 1995.

\bibitem{hopfield1982}
J.~J. Hopfield, ``Neural networks and physical systems with emergent collective
  computational abilities,'' \emph{Proceedings of the National Academy of
  Sciences}, vol.~79, no.~8, pp. 2554--2558, 1982.

\bibitem{jordan1986}
M.~I. Jordan, ``Serial order: {A} parallel, distributed processing approach,''
  Institute for Cognitive Science, University of California, San Diego, Tech.
  Rep. 8604, 1986.

\bibitem{elman1990}
J.~L. Elman, ``Finding structure in time,'' \emph{Cognitive Science}, vol.~14,
  no.~2, pp. 179--211, 1990.

\bibitem{werbos1990}
P.~J. Werbos, ``Backpropagation through time: what it does and how to do it,''
  \emph{Proceedings of the IEEE}, vol.~78, no.~10, pp. 1550--1560, 1990.

\bibitem{williams1989}
R.~J. Williams and D.~Zipser, ``A learning algorithm for continually running
  fully recurrent neural networks,'' \emph{Neural Computation}, vol.~1, no.~2,
  pp. 270--280, 1989.

\bibitem{bengio1994}
Y.~Bengio, P.~Simard, and P.~Frasconi, ``Learning long-term dependencies with
  gradient descent is difficult,'' \emph{IEEE Transactions on Neural Networks},
  vol.~5, no.~2, pp. 157--166, 1994.

\bibitem{martens2011}
J.~Martens and I.~Sutskever, ``Learning recurrent neural networks with
  hessian-free optimization,'' in \emph{Proceedings of the 28th International
  Conference on Machine Learning}, 2011, pp. 1033--1040.

\bibitem{schmidhuber2007}
J.~Schmidhuber, D.~Wierstra, M.~Gagliolo, and F.~Gomez, ``Training recurrent
  networks by {E}volino,'' \emph{Neural Computation}, vol.~19, no.~3, pp.
  757--779, 2007.

\bibitem{gomez2008}
F.~Gomez, J.~Schmidhuber, and R.~Miikkulainen, ``Accelerated neural evolution
  through cooperatively coevolved synapses,'' \emph{Journal of Machine Learning
  Research}, vol.~9, pp. 937--965, 2008.

\bibitem{hochreiter1997}
S.~Hochreiter and J.~Schmidhuber, ``Long short-term memory,'' \emph{Neural
  Computation}, vol.~9, no.~8, pp. 1735--1780, 1997.

\bibitem{cho2014}
K.~Cho, B.~van Merrienboer, C.~Gulcehre, D.~Bahdanau, F.~Bougares, H.~Schwenk,
  and Y.~Bengio, ``Learning phrase representations using {RNN}
  encoder{--}decoder for statistical machine translation,'' in
  \emph{Proceedings of the 2014 Conference on Empirical Methods in Natural
  Language Processing ({EMNLP})}.\hskip 1em plus 0.5em minus 0.4em\relax Doha,
  Qatar: Association for Computational Linguistics, 2014, pp. 1724--1734.

\bibitem{lukosevicius2009}
M.~Lukoševičius and H.~Jaeger, ``Reservoir computing approaches to recurrent
  neural network training,'' \emph{Computer Science Review}, vol.~3, pp.
  127--149, 2009.

\bibitem{jaeger2001}
H.~Jaeger, ``The “echo state” approach to analysing and training recurrent
  neural networks - with an erratum note,'' \emph{GMD Report 148, GMD - German
  National Research Institute for Computer Science}, 2001.

\bibitem{maass2002}
W.~Maass, T.~Natschl{\"a}ger, and H.~Markram, ``Real-time computing without
  stable states: a new framework for neural computation based on
  perturbations,'' \emph{Neural Computation}, vol.~14, no.~11, pp. 2531--2560,
  2002.

\bibitem{rabinovich2008}
M.~Rabinovich, R.~Huerta, and G.~Laurent, ``Transient dynamics for neural
  processing,'' \emph{Science}, vol. 321, no. 5885, pp. 48--50, 2008.

\bibitem{cover1965}
T.~M. Cover, ``Geometrical and statistical properties of systems of linear
  inequalities with applications in pattern recognition,'' \emph{IEEE
  Transactions on Electronic Computers}, vol.~14, no.~3, pp. 326--334, 1965.

\bibitem{ozturk2007}
M.~C. Ozturk, D.~Xu, and J.~C. Pr{\'\i}ncipe, ``Analysis and design of echo
  state networks,'' \emph{Neural Computation}, vol.~19, no.~1, pp. 111--138,
  2007.

\bibitem{tanaka2019recent}
G.~Tanaka, T.~Yamane, J.~B. H{\'e}roux, R.~Nakane, N.~Kanazawa, S.~Takeda,
  H.~Numata, D.~Nakano, and A.~Hirose, ``Recent advances in physical reservoir
  computing: A review,'' \emph{Neural Networks}, vol. 115, pp. 100--123, 2019.

\bibitem{brunner2013}
D.~Brunner, M.~C. Soriano, C.~R. Mirasso, and I.~Fischer, ``Parallel photonic
  information processing at gigabyte per second data rates using transient
  states,'' \emph{Nature Communications}, vol.~4, p. 1364, 2013.

\bibitem{larger2012}
L.~Larger, M.~C. Soriano, D.~Brunner, L.~Appeltant, J.~M. Gutierrez,
  L.~Pesquera, C.~R. Mirasso, and I.~Fischer, ``Photonic information processing
  beyond turing: an optoelectronic implementation of reservoir computing,''
  \emph{Optics Express}, vol.~20, no.~3, pp. 3241--3249, 2012.

\bibitem{paquot2012}
Y.~Paquot, F.~Duport, A.~Smerieri, J.~Dambre, B.~Schrauwen, M.~Haelterman, and
  S.~Massar, ``Optoelectronic reservoir computing,'' \emph{Scientific Reports},
  vol.~2, p. 287, 2012.

\bibitem{rodan2011}
A.~Rodan and P.~Tino, ``Minimum complexity echo state network,'' \emph{IEEE
  Transactions on Neural Networks}, vol.~22, no.~1, pp. 131--144, 2011.

\bibitem{appeltant2011}
L.~Appeltant, M.~C. Soriano, G.~Van~der Sande, J.~Danckaert, S.~Massar,
  J.~Dambre, B.~Schrauwen, C.~R. Mirasso, and I.~Fischer, ``Information
  processing using a single dynamical node as complex system,'' \emph{Nature
  Communications}, vol.~2, p. 468, 2011.

\bibitem{strauss2012}
T.~Strauss, W.~Wustlich, and R.~Labahn, ``Design strategies for weight matrices
  of echo state networks,'' \emph{Neural Computation}, vol.~24, no.~12, pp.
  3246--3276, 2012.

\bibitem{appeltant2014}
L.~Appeltant, G.~Van~der Sande, J.~Danckaert, and I.~Fischer, ``Constructing
  optimized binary masks for reservoir computing with delay systems,''
  \emph{Scientific Reports}, vol.~4, p. 3629, 2014.

\bibitem{triesch2005}
J.~Triesch, ``A gradient rule for the plasticity of a neuron’s intrinsic
  excitability,'' in \emph{International Conference on Artificial Neural
  Networks}.\hskip 1em plus 0.5em minus 0.4em\relax Springer, 2005, pp. 65--70.

\bibitem{langton1990}
C.~G. Langton, ``Computation at the edge of chaos: Phase transitions and
  emergent computation,'' \emph{Physica D: Nonlinear Phenomena}, vol.~42,
  no.~1, pp. 12--37, 1990.

\bibitem{bak1988}
P.~Bak, C.~Tang, and K.~Wiesenfeld, ``Self-organized criticality,''
  \emph{Physical Review A}, vol.~38, pp. 364--374, 1988.

\bibitem{kauffman1993}
S.~A. Kauffman, \emph{The Origins of Order: Self-organization and selection in
  evolution}.\hskip 1em plus 0.5em minus 0.4em\relax Oxford University Press,
  1993.

\bibitem{busing2010}
L.~Büsing, B.~Schrauwen, and R.~Legenstein, ``Connectivity, dynamics, and
  memory in reservoir computing with binary and analog neurons,'' \emph{Neural
  Computation}, vol.~22, no.~5, pp. 1272--1311, 2010.

\bibitem{bertschinger2004}
N.~Bertschinger and T.~Natschläger, ``Real-time computation at the edge of
  chaos in recurrent neural networks,'' \emph{Neural Computation}, vol.~16,
  no.~7, pp. 1413--1436, 2004.

\bibitem{snyder2013}
D.~Snyder, A.~Goudarzi, and C.~Teuscher, ``Computational capabilities of random
  automata networks for reservoir computing,'' \emph{Physical Review E},
  vol.~87, p. 042808, 2013.

\bibitem{jaeger2004}
H.~Jaeger and H.~Haas, ``Harnessing nonlinearity: Predicting chaotic systems
  and saving energy in wireless communication,'' \emph{Science}, vol. 304, no.
  5667, pp. 78--80, 2004.

\bibitem{strogatz1994}
S.~H. Strogatz, \emph{Nonlinear dynamics and chaos, with applications to
  physics, biology, chemistry, and engineering}.\hskip 1em plus 0.5em minus
  0.4em\relax Westview Press, 1994.

\bibitem{verstraeten2007}
D.~Verstraeten, B.~Schrauwen, M.~D’Haene, and D.~Stroobandt, ``An
  experimental unification of reservoir computing methods,'' \emph{Neural
  Networks}, vol.~20, no.~3, pp. 391--403, 2007.

\bibitem{verstraeten2009}
D.~Verstraeten and B.~Schrauwen, ``On the quantification of dynamics in
  reservoir computing,'' in \emph{Artificial Neural Networks -- ICANN
  2009}.\hskip 1em plus 0.5em minus 0.4em\relax Springer Berlin Heidelberg,
  2009, pp. 985--994.

\bibitem{boedecker2012}
J.~Boedecker, O.~Obst, J.~T. Lizier, N.~M. Mayer, and M.~Asada, ``Information
  processing in echo state networks at the edge of chaos,'' \emph{Theory in
  Biosciences}, vol. 131, no.~3, pp. 205--213, 2012.

\bibitem{livi2018}
L.~{Livi}, F.~M. {Bianchi}, and C.~{Alippi}, ``Determination of the edge of
  criticality in echo state networks through {F}isher information
  maximization,'' \emph{IEEE Transactions on Neural Networks and Learning
  Systems}, vol.~29, no.~3, pp. 706--717, 2018.

\bibitem{legenstein2005}
R.~Legenstein and W.~Maass, ``What makes a dynamical system computationally
  powerful?'' in \emph{New Directions in Statistical Signal Processing: From
  Systems to Brain}, S.~Haykin, J.~Príncipe, T.~Sejnowski, and J.~McWhirter,
  Eds.\hskip 1em plus 0.5em minus 0.4em\relax MIT Press, 2007, pp. 127--154.

\bibitem{bailey1996}
B.~A. Bailey, ``Local {L}yapunov exponents: predictability depends on where you
  are,'' in \emph{Nonlinear Dynamics and Economics}, vol.~10.\hskip 1em plus
  0.5em minus 0.4em\relax Cambridge University Press, 1996, pp. 345--359.

\bibitem{bianchi2018}
F.~M. Bianchi, L.~Livi, and C.~Alippi, ``Investigating echo-state networks
  dynamics by means of recurrence analysis,'' \emph{IEEE Transactions on Neural
  Networks and Learning Systems}, vol.~29, no.~2, pp. 427--439, 2018.

\bibitem{white2004}
O.~L. White, D.~D. Lee, and H.~Sompolinsky, ``Short-term memory in orthogonal
  neural networks,'' \emph{Phys. Rev. Lett.}, vol.~92, p. 148102, 2004.

\bibitem{jaeger2002}
H.~Jaeger, ``Short term memory in echo state networks,'' \emph{GMD Report 152,
  GMD - German National Research Institute for Computer Science}, 2002.

\bibitem{verstraeten2010memory}
D.~Verstraeten, J.~Dambre, X.~Dutoit, and B.~Schrauwen, ``Memory versus
  non-linearity in reservoirs,'' in \emph{The 2010 international joint
  conference on neural networks (IJCNN)}.\hskip 1em plus 0.5em minus
  0.4em\relax IEEE, 2010, pp. 1--8.

\bibitem{dambre2012}
J.~Dambre, D.~Verstraeten, B.~Schrauwen, and S.~Massar, ``Information
  processing capacity of dynamical systems,'' \emph{Scientific Reports},
  vol.~2, p. 514, 2012.

\bibitem{digregorio2018}
E.~Di~Gregorio, C.~Gallicchio, and A.~Micheli, ``Combining memory and
  non-linearity in echo state networks,'' in \emph{Artificial Neural Networks -
  ICANN 2018}, 2018, pp. 556--566.

\bibitem{inubushi2017}
M.~Inubushi and K.~Yoshimura, ``Reservoir computing beyond memory-nonlinearity
  trade-off,'' \emph{Scientific Reports}, vol.~7, no.~1, p. 10199, 2017.

\bibitem{atiya2000}
A.~F. {Atiya} and A.~G. {Parlos}, ``New results on recurrent network training:
  unifying the algorithms and accelerating convergence,'' \emph{IEEE
  Transactions on Neural Networks}, vol.~11, no.~3, pp. 697--709, 2000.

\bibitem{mackey1977}
M.~Mackey and L.~Glass, ``Oscillation and chaos in physiological control
  systems,'' \emph{Science}, vol. 197, no. 4300, pp. 287--289, 1977.

\bibitem{gallicchio2017desn}
C.~Gallicchio, A.~Micheli, and L.~Pedrelli, ``Deep reservoir computing: A
  critical experimental analysis,'' \emph{Neurocomputing}, vol. 268, pp.
  87--99, 2017.

\bibitem{gallicchio2020fast}
C.~Gallicchio and A.~Micheli, ``Fast and deep graph neural networks,'' in
  \emph{Proceedings of the AAAI Conference on Artificial Intelligence},
  vol.~34, no.~04, 2020, pp. 3898--3905.

\bibitem{lazar2007}
A.~Lazar, G.~Pipa, and J.~Triesch, ``Fading memory and time series prediction
  in recurrent networks with different forms of plasticity,'' \emph{Neural
  Networks}, vol.~20, no.~3, pp. 312 -- 322, 2007.

\end{thebibliography}

\end{document}